\documentclass[sigconf]{acmart}
\usepackage{import}
\usepackage{algorithmic}
\usepackage{algorithm}

\usepackage{textcomp}
\usepackage{multirow}
\usepackage{booktabs}
\usepackage{amsmath,xparse,mleftright}
\usepackage{balance}

\AtBeginDocument{%
  \providecommand\BibTeX{{%
    \normalfont B\kern-0.5em{\scshape i\kern-0.25em b}\kern-0.8em\TeX}}}

\copyrightyear{2022}
\acmYear{2022}
\setcopyright{acmcopyright}\acmConference[MM '22]{Proceedings of the 30th ACM
International Conference on Multimedia}{October 10--14, 2022}{Lisboa, Portugal}
\acmBooktitle{Proceedings of the 30th ACM International Conference on Multimedia
(MM '22), October 10--14, 2022, Lisboa, Portugal}
\acmPrice{15.00}
\acmDOI{10.1145/3503161.3548344}
\acmISBN{978-1-4503-9203-7/22/10}

\begin{document}

\title{RCRN: Real-world Character Image Restoration Network via Skeleton Extraction}


\author{Daqian Shi}
\email{daqian.shi@unitn.it}
\orcid{0000-0003-2183-1957}
\affiliation{%
  \institution{College of Computer Science and Technology, Jilin University}
  \country{}
}
\additionalaffiliation{%
  \institution{DISI, University of Trento}
  \country{}
}

\author{Xiaolei Diao}
\email{xiaolei.diao@unitn.it}
\orcid{0000-0002-3269-8103}
\affiliation{%
  \institution{DISI, University of Trento}
  \country{}
}

\author{Hao Tang}
\email{hao.tang@vision.ee.ethz.ch}
\affiliation{%
  \institution{CVL, ETH Zurich}
  \country{}
}

\author{Xiaomin Li}
\email{xmli19@mails.jlu.edu.cn}
\orcid{0000-0001-7202-6865}
\affiliation{%
  \institution{School of Artificial Intelligence, Jilin University}
  \country{}
}

\author{Hao Xing}
\email{xinghao19@mails.jlu.edu.cn}
\orcid{0000-0002-4254-5282}
\affiliation{%
  \institution{School of Artificial Intelligence, Jilin University}
  \country{}
}

\author{Hao Xu}
\email{xuhao@jlu.edu.cn}
\orcid{0000-0001-8474-0767}
\affiliation{%
  \institution{College of Computer Science and Technology, Jilin University}
  \country{}
}
\additionalaffiliation{%
  \institution{Symbol Computation and Knowledge Engineer of Ministry of Education, Jilin University}
  \country{}
}
\authornote{Corresponding author}


\renewcommand{\shortauthors}{Daqian Shi et al.}



\begin{abstract}
Constructing high-quality character image datasets is challenging because real-world images are often affected by image degradation. There are limitations when applying current image restoration methods to such real-world character images, since (i) the categories of noise in character images are different from those in general images; (ii) real-world character images usually contain more complex image degradation, e.g., mixed noise at different noise levels. To address these problems, we propose a real-world character restoration network (RCRN) to effectively restore degraded character images, where character skeleton information and scale-ensemble feature extraction are utilized to obtain better restoration performance. The proposed method consists of a skeleton extractor (SENet) and a character image restorer (CiRNet). SENet aims to preserve the structural consistency of the character and normalize complex noise. Then, CiRNet reconstructs clean images from degraded character images and their skeletons. Due to the lack of benchmarks for real-world character image restoration, we constructed a dataset containing 1,606 character images with real-world degradation to evaluate the validity of the proposed method. The experimental results demonstrate that RCRN outperforms state-of-the-art methods quantitatively and qualitatively.
\end{abstract}




\maketitle

\section{Introduction}
\begin{figure}[!t]
\setlength{\abovecaptionskip}{5pt}%
\setlength{\belowcaptionskip}{0pt}%
	\centering
	\includegraphics[width=0.9\linewidth]{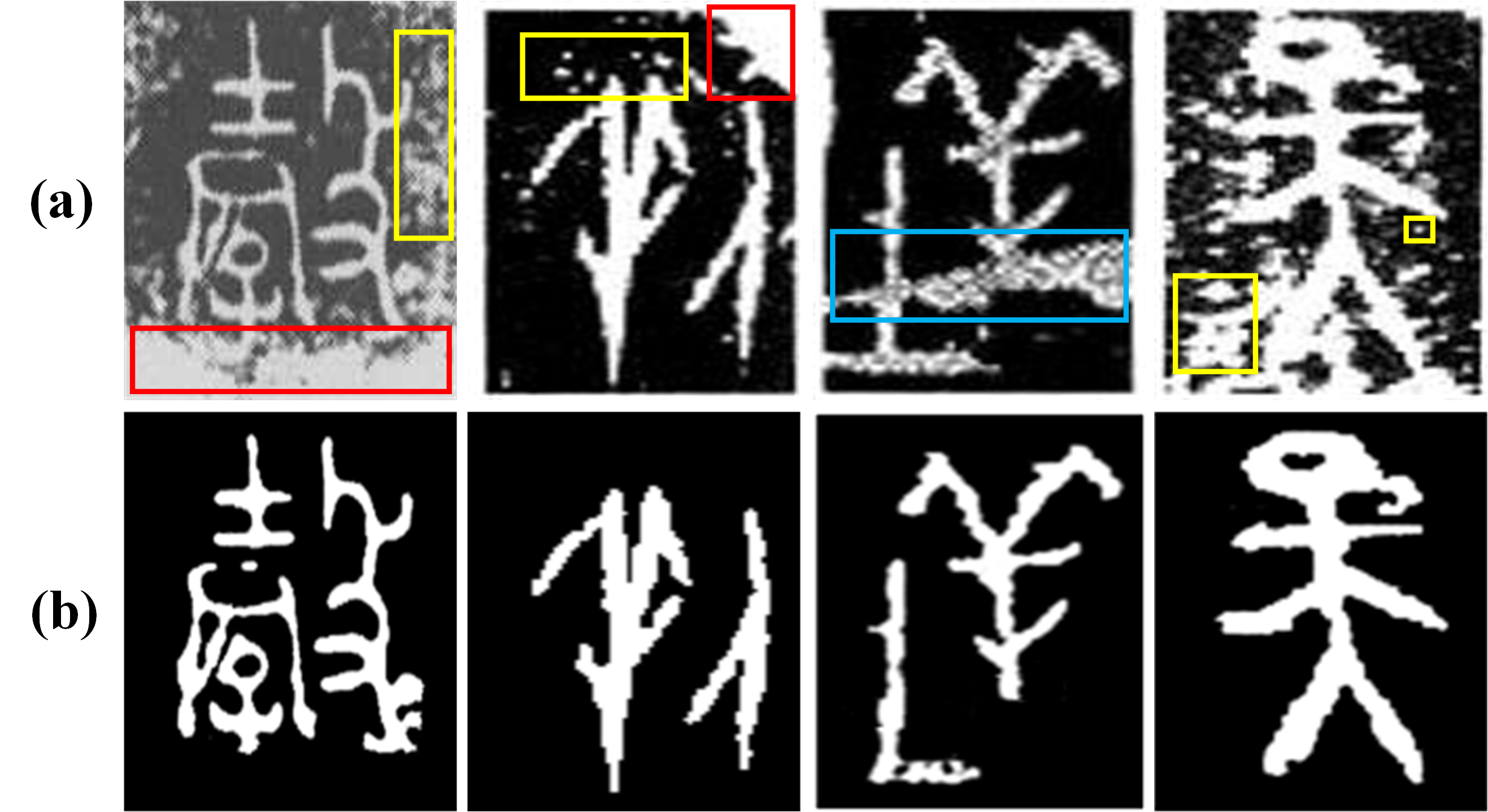}
	\caption{Demonstration of character image restoration for historical characters. (a) real-world character images with complex degradation, where different categories of degradation are discriminated by coloured boxes; (b) restoration results by our method, where most of the degradation is removed and characters are restored. \label{fig:1}}
\end{figure}

Recently, character image datasets have been widely applied and studied due to the increase in research topics such as optical character recognition \cite{6}. Generally, character image is a broad concept in the research community, which includes different datasets (e.g., documents, historical and street-view texts) in various languages (e.g., English, Chinese) and fonts (e.g., handwritten and printed texts). However, real-world character image datasets are often unsatisfactory to the requirements in practice, where image degradation is one of the common issues. For example, historical Chinese character images \cite{47}, which are high-valuable and irreplaceable, often have a large amount of natural noise. Such degraded character images not only challenge the management of the digital library but also mislead human understanding and interpretation \cite{7}. Therefore, it is necessary to investigate how to effectively restore real-world degraded character images.

Recent studies \cite{8, 23, 24} pointed out that real-world character images, e.g., degraded document images, often have specific categories of degradation like ink smear, erosion noise, interfering patterns, and excavation damage. Unlike synthetic noise models, such as Gaussian noise \cite{2} and salt-and-pepper noise \cite{3}, it is more complex to model real-world degradation using probability density functions \cite{39}. Mixed noise and different noise levels probably appear in the same dataset, making the image restoration more challenging. Figure~\ref{fig:1}(a) presents examples of degraded character images, where different noise categories are highlighted in boxes with different colours.
The first two images both contain mixed erosion (yellow) and broken edge noise (red), the third image contains cracking noise (blue), and the fourth image contains erosion noise (yellow) at different noise levels. In this paper, we aim to address this problem of low-level visibility degradation. Given a real-world degraded character image, our goal is to remove the various noise and restore the character, as shown in Figure~\ref{fig:1}(b).

Inspired by studies on general image denoising, early research on character image restoration mainly includes spatial pixel feature denoising \cite{9, 10}  and variable domain denoising \cite{36, 13}. These methods perform poorly in practice since they are designed to remove synthetic noise \cite{8}. Some dedicated methods for specific degradation are introduced to solve this problem, such as continuous area calculation to remove erosion noise \cite{22} and multispectral analysis to restore uneven background \cite{33}. However, one prerequisite for applying dedicated methods is to know the noise category and level in advance, which is difficult to satisfy in a real-world image restoration scenario.

Recently, deep neural networks have been applied for character image restoration. Methods based on denoising convolutional neural networks (DnCNNs) \cite{16} have achieved progress on character image restoration tasks \cite{17,18}. Adversarial learning-based methods have also been introduced, e.g., an adversarial autoencoder \cite{15} is introduced for removing salt-and-pepper and Gaussian noise from historical document images. However, these studies mainly focus on presenting and removing synthetic noise, since synthetic noise is easier to learn and remove. Therefore, it is necessary to develop general methods for real-world character image restoration.

The above observations bring two main issues into focus. First, the categories of noise in character images are different from those in general images. Second, real-world character images usually contain more complex image degradation, e.g., mixed noise at different noise levels. To fix both issues, in this paper, we design an end-to-end generative adversarial network (GAN)-based framework for real-world degraded character image restoration, called RCRN. Unlike existing methods, RCRN intends to utilize character skeleton information and scale-ensemble feature extraction to obtain better image restoration performance.
RCRN consists of a skeleton extractor (SENet) and a character image restorer (CiRNet). SENet aims to extract stable skeletons from the degraded character image by applying a skeleton-ensemble strategy. It can preserve the structural consistency of the target character and normalize complex noise. We proposed CiRNet to reconstruct clean character images from the degraded images and their skeletons, where a scale-ensemble-based network and a novel skeleton loss function are introduced to better deal with mixed noise. 

Overall, our contributions can be summarized as follows:

\begin{itemize}
    \item We propose a novel GAN-based real-world character image restoration framework, i.e., RCRN. It can effectively handle complex image degradation and specific noise categories in real-world character images. 
    \item We utilize character skeleton information to reconstruct the character image and recover the degradation while preserving its structural consistency. The skeleton-ensemble strategy is applied for extracting stable skeletons from noisy images. Moreover, we propose a skeleton loss function to enhance model training.
    \item We construct a new character image dataset, where real-world degraded images are included. We compare the proposed RCRN with state-of-the-art methods, and experimental results demonstrate the superiority of our method, in particular for character images with real-world degradation.
\end{itemize}

\section{Related Work}
\subsection{GAN-Based Image Restoration}
Recently, GANs \cite{27} are gradually applied for image restoration. Some GAN-based methods attempt to model image degradation by learning from clean-noisy image pairs \cite{5}. Inspired by image-to-image translation, an attentive GAN \cite{30} is proposed to remove raindrops from degraded images by injecting visual attention into the networks. By applying capsule networks, \cite{52} proposes an adversarial learning framework that introduces loss terms spanning three domains for median filtered image restoration. \cite{51} proposes an image deraining method that includes a depth-guided GAN backbone for estimating rain streaks and transmission. The generated rain will be removed from the image in the second stage.

Due to the lack of paired training data for real-world images, GCBD \cite{28} introduces a two-step training method, where a GAN-based noise estimator is trained to generate image pairs for training the denoiser. A physics-guided generative adversarial framework \cite{49} is proposed for image restoration tasks, including image deblurring, dehazing and deraining. In the medical image processing area, GAN-based denoisers \cite{32, 35} are applied to remove granular speckle noise in optical coherence tomography. \cite{50} proposes a novel method based on 3D conditional generative adversarial networks to estimate the high-quality full-dose positron emission tomography images from low-dose images. The success of the above studies inspires us to develop a general method for restoring real-world character images based on GAN.

\subsection{Character Image Restoration}
Character image denoising has attracted significant attention since the last century. Learning from general denoising methods, people apply filter-based \cite{9}, total variation-based \cite{13}, and diffusion-based \cite{12} methods to remove synthetic noise in early research. A comparative study \cite{8} evaluates several common techniques for Chinese character image restoration, such as non-local means (NLM)~\cite{21}. Using the divide-and-conquer strategy, an ensemble-based method \cite{20} stacks region NLM filter into Markovian segmentation to remove Gaussian noise and smooth document images. 

Considering that there are additional noise categories in real-world character images compared with general images \cite{8}. Researchers have designed some dedicated restoration methods for the corresponding noise categories. 
For instance, \cite{22} develops an oracle image restoration strategy to distinguish erosion noise from target characters through fractal geometry analysis and character area calculation. \cite{23} proposes an ensemble denoising method to restore uneven background in degraded historical document images. Dedicated character denoisers \cite{10,13} are proposed based on KSVD dictionary learning and character strokes to remove ant-like interfering patterns appearing in historical character rubbings.

Besides, deep learning models, such as DnCNN, have been applied to restore character images under the blind noise reduction scenario. As an improvement, \cite{17} proposes a method that combined DnCNN and median filtering to remove additive white Gaussian noise (AWGN) at  different levels from ancient character images. Noise2Same \cite{18} introduces a self-supervised denoising framework for degraded character images, which can model and remove Gaussian noise without inputting additional information. An adaptive image binarization method based on a variational model was designed by \cite{26}, which is applied for unknown level AWGN denoising in document images. A GAN-based character image denoiser \cite{14} can generate clean Chinese calligraphic images by learning from degraded images with mixed Gaussian and salt-and-pepper noise. 

Learning from the above methods, we found that the dedicated method designed for the corresponding category of noise is limited in practice since the user needs to know the category and level of noise in advance. While general restoration methods mainly focus on removing synthetic noise and perform poorly on real-world character images. To solve these problems, in this paper, we propose a general end-to-end image restoration method that can apply to real-world character images.

\section{Intuitive Discussion}
The goal of the character image restoration task is to recover a clean image $X \in \mathbb{R}^{H\times W}$ from a noisy observation $Y \in \mathbb{R}^{H\times W}$. In this section, we first model real-world character image degradation to clarify the task in this paper. Then, we provide some intuitive discussions on potential improvements to deal with complex degradation and better image restoration performance, driving the motivation of this paper.

\subsection{Problem Formulation}

Existing image restoration methods, e.g., DnCNN \cite{16}, generally assume a noisy observation $Y$ follows an image degradation model $Y = X + N$, where $N$ is a specific category of synthetic noise distribution like AWGN. These methods intend to estimate and remove the noise $N$ from $Y$ to restore images. However, they do not perform well on real-world degraded images, since the degradation model for real-world character images is more complex and different from the synthetic noise above \cite{5}. It contains additive noise independent of the character, e.g., ink smear, cracking damage, and background noise. Moreover, considering the mixed noise categories and the uncertain noise level, we model the real-world degraded character image $Y$ as:   
\begin{equation}
    Y = X + \sum f(N_i),
\label{equ:1}
\end{equation}
where $N_i$ refers to the distribution of a specific noise category; $f(\cdot)$ presents the distribution function for uncertain noise level, $f(N_i) {\sim} \mathcal{N} (0, \sigma^2(N_i))$; and the sum of $f(N_i)$ represents the distribution of mixed noise. 

Furthermore, real-world images include signal-dependent noise, e.g., in-camera processing and image compression noise, which further increase the complexity of the degradation \cite{39}. We also consider such noise in the degradation model as:
\begin{equation}
    Y = W( X + \sum f(N_i) ),
\label{equ:2}
\end{equation}
where $W(\cdot)$ indicates the distribution function for signal-dependent noise. According to the proposed degradation model (Eq.~\eqref{equ:2}) for real-world degraded images, we can find it is costly to model real-world degradation since its complexity. And the key conditions (noise level $f(\cdot)$, noise category $N_i$ and signal-dependent noise function $W(\cdot)$) are inaccessible in advance under the real-world scenario (e.g., blind-denoising). Thus, the intuition is to restore such images by reconstructing the character part rather than estimating and removing degradation from the image. In this paper, our goal is to reconstruct a clean character image $X$ from a noisy observation $Y$, without giving the remaining conditions.

\subsection{Observations and Motivations}

\begin{figure}[!t]
	\centering
	\setlength{\abovecaptionskip}{2pt}%
    \setlength{\belowcaptionskip}{0pt}%
	\includegraphics[width=0.78\linewidth]{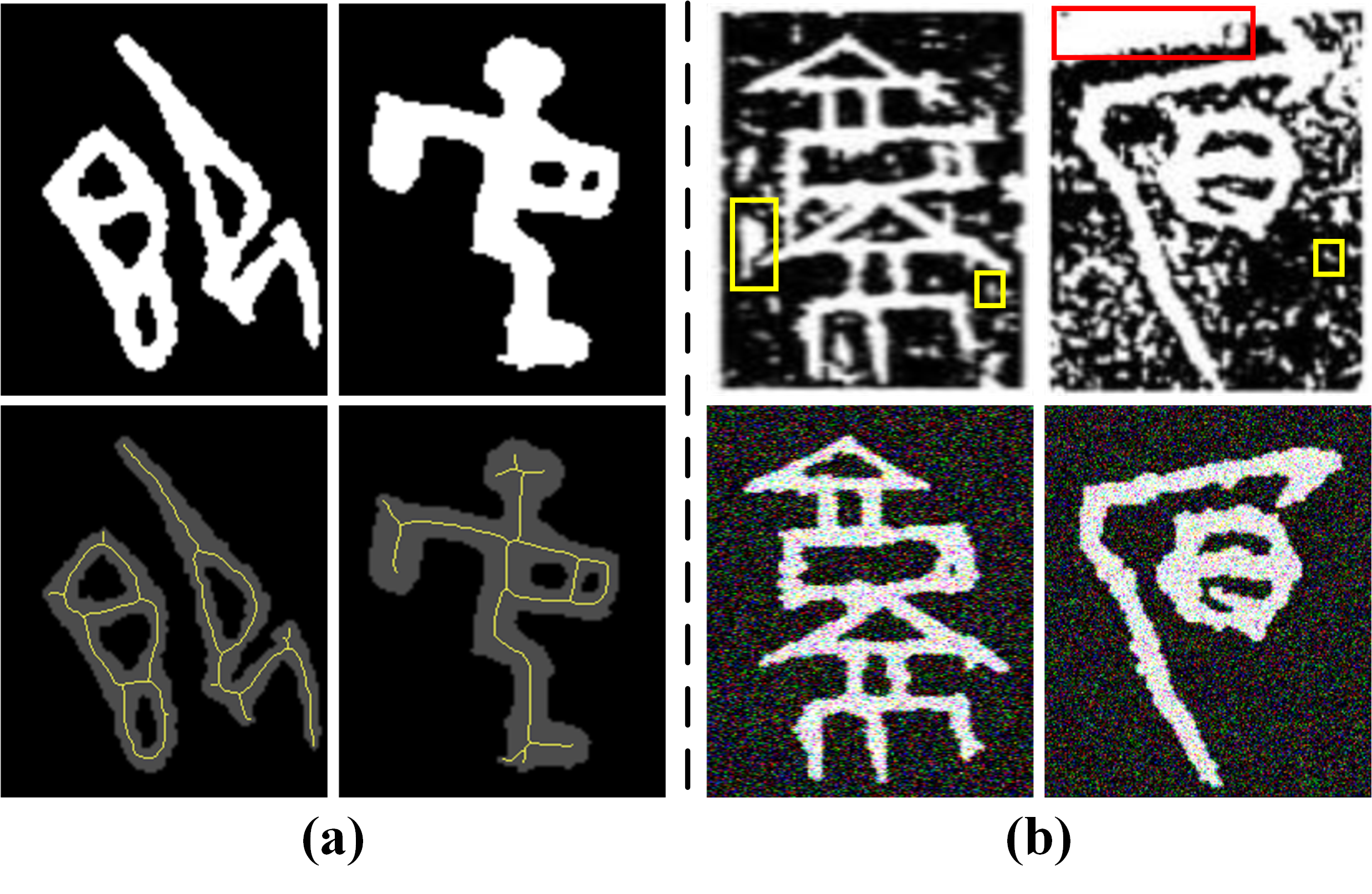}
	\caption{Examples of our observations. (a) skeletonization demonstration; (b) Comparison between different categories of degradation.	\label{fig:new1}}
\end{figure}

\noindent\textbf{Skeletons for Obtaining Character Structure Information.}
The character structure, which comes from the combination of character strokes, is the most crucial feature to maintain the semantic coherence of the character \cite{7}. In real-world character image restoration, we need to reconstruct the clean character from complex degradation without destroying the structure of the character; otherwise, the meaning of the character will change. People suppose that the skeleton can be used to capture the structural information of a character \cite{34}. Thus, skeletonization methods are applied to present character structures and structural features. As an example, Figure \ref{fig:new1} (a) demonstrates several results of character image skeletonization, where the first row presents original characters and the second row highlights the skeletons (in yellow) of these characters.

\begin{figure*}[!t]
	\centering
	\includegraphics[width=0.99\linewidth]{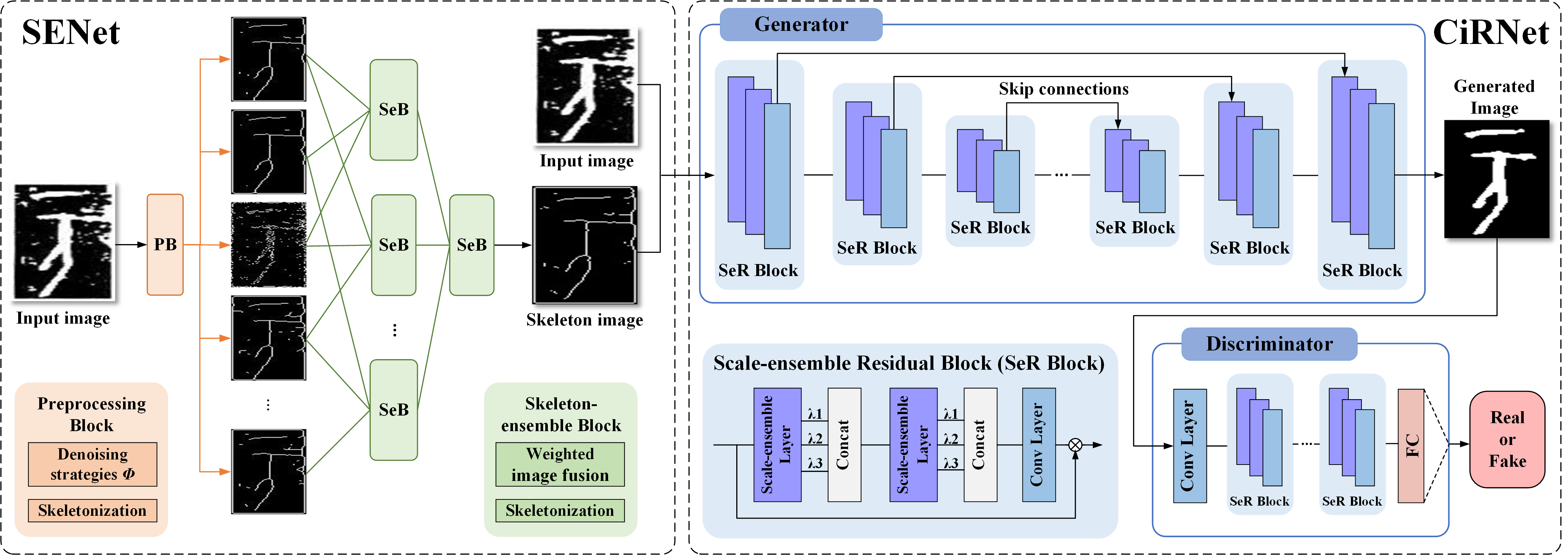}
	\caption{The overall structure of RCRN. We use different colors to represent different modules in each network. SENet (left) extracts skeletons from input noisy character images and CiRNet (right) generates clean character images as the final output.
	\label{fig:2}}
\end{figure*}

\noindent\textbf{Difference between Noise Categories.}
We have discussed the challenge of modeling and recognizing real-world degradation compared to common synthetic degradation. We present several concrete examples in Figure \ref{fig:new1} (b), where the first row shows the real-world degradation and the second shows synthetic degradation (Gaussian noise with variance $\sigma = 10$). We can distinguish synthetic degradation from character, since the scale and texture between degradation and character are visibly different. On the contrary, it is not easy for real-world degradation cases, especially for images with complex degradation (e.g., mixed noise). As shown in the first row of Figure \ref{fig:new1} (b), degradation with different scales (highlighted in boxes) challenges image restoration methods.

\noindent\textbf{Motivation and Challenges.}
As a result, we obtain two crucial ideas from the observations we discussed above for improving real-world character image restoration. Firstly, we find that skeletons can capture character structure information, which will help to improve the performance of character image reconstruction. We observe that current skeletonization methods are not performing well in our case since they are not designed for images with complex degradation. Thus, we propose SENet to extract stable and complete skeletons from degraded images by exploiting a skeleton ensemble strategy. Moreover, applying skeletonization can also normalize complex degradation to the same level, which will help alleviate the complexity of noise recognition and location. 

Second, we observe that real-world degraded images are more difficult to restore, where a powerful feature extraction ability is required to deal with the degradation with different scales at the same step. Thus, we propose CiRNet to effectively restore real-world character images by considering multiple receptive fields in the network, which will help to recognize and locate noise in complex cases.

\section{The Proposed RCRN}

\subsection{SENet}
The proposed RCRN consists of SENet and CiRNet, as shown in Figure~\ref{fig:2}. We discussed that character structure is a critical feature that can be captured by character skeletons for better restoration performance. However, applying existing skeletonization methods to degraded character images generally cannot obtain reasonable skeletons, since these methods are dedicated to clean images and vulnerable to degradation. Thus, we introduce SENet, a general ensemble-based skeletonization network, to minimize the effect of diverse noise and obtain reasonable skeletons of character images. 

The structure of SENet is shown on the left of Figure~\ref{fig:2}, where we introduce random-selection pre-processing and skeleton-ensemble strategy to deal with the uncertain effects of skeletonization under complex degradation. Firstly, in the random-selection pre-processing block, we define a processing pool $P = \{P_i | i = 1,2,...,n\}$ to store several physics-guided methods for image transformation and denoising, including adaptive threshold/histogram equalization, gamma transformation, and bilateral/Gaussian/Median filters with three different kernel sizes. According to pool $P$, we define a denoising function $\phi$ as a sequence of processing. Thus, we have:  
\begin{equation}
    \phi(Y) = P_i(P_j(P_k(Y))),
\label{equ:3.1}
\end{equation}
where $Y$ is the noisy character image; $P_i, P_j$, and $P_k$ are selected from $P$ for transforming $Y$ into diverse spatial matrices. Note that a set of denoising functions $\Phi$ will be generated, where $\phi \in \Phi$. These functions will help to improve the robustness of the skeletonization result by considering different transformations and scales. We stack skeletonization process after each denoising function, as $sk = SK(\phi(Y))$. Thus, by inputting a noisy character image $Y$, we will collect a set of skeletons $S^1$ corresponding to denoising functions from the pre-processing block, where we have $sk \in S^1$.

Then, we apply skeleton-ensemble blocks for extracting the common skeleton features from $S^1$ by fusing multiple skeletons, which also aims to improve the robustness of skeletons. For each of the skeleton-ensemble blocks, we apply weighted image fusion on $n$ randomly selected skeletons from $S^1$. We skeletonize the fusion results as a new set of skeletons $S^2$. We repeatedly apply such ensemble blocks until the output $S$ becomes a single image. As shown in Figure~\ref{fig:2}, the skeleton image $S$ will be concatenated with the original noisy character image $Y$ as the input of CiRNet for providing the character structural information.   

\begin{figure}[!t]
	\centering
	\setlength{\abovecaptionskip}{2pt}%
    \setlength{\belowcaptionskip}{2pt}%
	\includegraphics[width=0.95\linewidth]{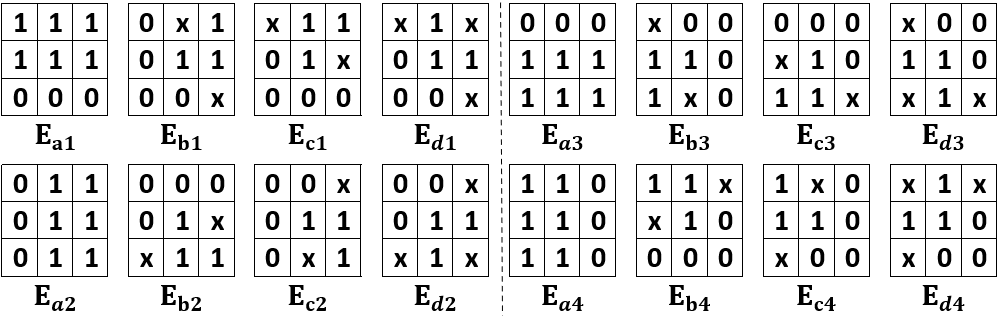}
	\caption{The set of erosion kernels we propose for character skeletonization, where ``0'', ``1'', and ``\textsc{x}'' indicate that the target pixel is white, black, and arbitrary pixels, respectively.
	\label{fig:3}}
\end{figure}

Additionally, we inspire by the mathematical morphology-based skeletonization methods \cite{37} which exploit a set of kernels to erode and dilate the binarized character images. Considering the features of character strokes, a group of erosion kernels is applied in this work, as shown in Figure~\ref{fig:3}, where $E$ refers to the group of erosion kernels, ``a'' to ``d'' refer to different kinds of kernels and ``1'' to ``4'' refer to the directions of the kernels. We also discard the dilatation kernels to make our skeletonization method more lightweight, aiming to speed up the calculation.

The main idea of the skeletonization method is to match the binarized character image $C$ with erosion kernels $E = \{E_{a1}, E_{b1}, ..., E_{d4}\}$. Suppose $C^{xy}$ is a $3 \times 3$ patch centered on pixel $C_{xy}$ and $e$ is one of the kernels in set $E$. We update the value of the pixel $C_{xy}$ according to the matching operation~$\otimes$, we have: 

\begin{equation}
    C^{xy} \otimes e = 1- \prod_{i=1}^3 \prod_{j=1}^3 C^{xy}_{ij} \cdot e_{ij},
\label{equ:3}
\end{equation}
where $i$, $j$ refer to the position of the pixel in each patch or kernel. Thus, the erosion process is defined as: 
\begin{equation}
    C^{xy}_{ij} \cdot e_{ij} = \left\{
\begin{tabular}{ll}
1,  & if $C^{xy}_{ij} = e_{ij} \lor C^{xy}_{ij} = \textsc{x}$\\
0,  & otherwise 
\end{tabular}\right.
\label{equ:4}
\end{equation}

According to Eq.~\eqref{equ:4}, we obtain image $sk_1$ by matching the character image $C$ with all erosion kernels and updating the value of every pixel for $C$. Then, we input $sk_i$ and apply the above processing repeatedly until the skeletonized image $sk_{i+1}$ no longer changes, where $i = \{1, 2, ..., n\}$. Finally, we obtain $sk_n = SK(C)$, where $sk_n$ is the single-pixel skeleton of the input character image~$C$.

\subsection{CiRNet}

CiRNet is designed to generate clean character images $X$ by learning from noisy-clean image pairs. We develop CiRNet based on cGAN, where a scale-ensemble residual (SeR) block is proposed to deal with complex degradation with different scales. CiRNet will gradually learn to reconstruct clean character images through adversarial training, and the output image ought to preserve the character's structural consistency. The overall structure of CiRNet is shown on the right of Figure~\ref{fig:2}. 

\noindent \textbf{Generator of CiRNet.} In CiRNet, the generator is the backbone to reconstruct clean character images $\hat{y}$. As shown in Figure~\ref{fig:2}, our generator is developed as a U-shape structure which mainly consists of SeR blocks, where skip connections are utilized to prevent blurred outputs. Each SeR block consists of scale-ensemble layers, concatenation processes $\rm Concat$ and a convolution layer. 

As we discussed that it is challenging to restore real-world images containing degradation in different scales. The purpose of building scale-ensemble layers is to enhance feature extraction and improve restoration performance by introducing receptive fields in different scales. We intend to build short-distance and also long-distance dependency on complex degradation. $\rm Concat$ will concatenate outputs in different scales $\lambda = \{\lambda_1, \lambda_2, \lambda_3\}$. Finally, we pass the output to a convolution layer and apply a residual connection as the result of the SeR block. We propose two kinds of SeR blocks SeR-T and SeR-R based on different scale-ensemble layers.

\noindent \textbf{SeR-T.} We intend to introduce the self-attention mechanism into SeR-T blocks for feature extraction. SeR-T blocks apply vision Transformers (ViT) layers \cite{48} as the scale-ensemble layer. For obtaining features in different scales, we identify non-overlapped regions as windows to utilize local self-attention based on grids of images. Thus, the size of windows determines the receptive field for applying local self-attention. Applying ViT layers with different local self-attention window sizes will extract features in corresponding scales. Assume $F$ is the input feature map, we have: 
\begin{equation}
    F^t = {\rm ViT}(F, \lambda*IS),
\label{equ:7}
\end{equation}
where $F^t$ refers to the output feature map; $\rm ViT(\cdot)$ refers to the processing by ViT layers; $IS$ presents the size of local self-attention windows and $\lambda{=} \{1, 0.5, 0.25\}$.

\noindent \textbf{SeR-C.} We also adapt convolutional layers for extracting features in different scales, i.e., SeR-C blocks. In this stage, we consider two main requirements: (i) building a large receptive field needs to capture long-distance dependence by large kernels; (ii) the increasing calculation cost when enlarging the kernel size of the convolutional layer. To fulfil these two requirements, we exploit dilated convolution \cite{44} in scale-ensemble layers, where we stack three dilated convolution layers with different dilatation rates $\lambda$ in parallel. Suppose $F$ is the input feature map, we have:
\begin{equation}
    F^d = {\rm DilatedConv}(F, \lambda),
\label{equ:6}
\end{equation}
where $F^d$ represents the feature map processed by dilated convolution layers $\rm DilatedConv(\cdot)$, $\lambda = \{1, 2, 3\}$.

\begin{figure}[!t]
	\centering
	\setlength{\abovecaptionskip}{2pt}%
    \setlength{\belowcaptionskip}{0pt}%
	\includegraphics[width=0.95\linewidth]{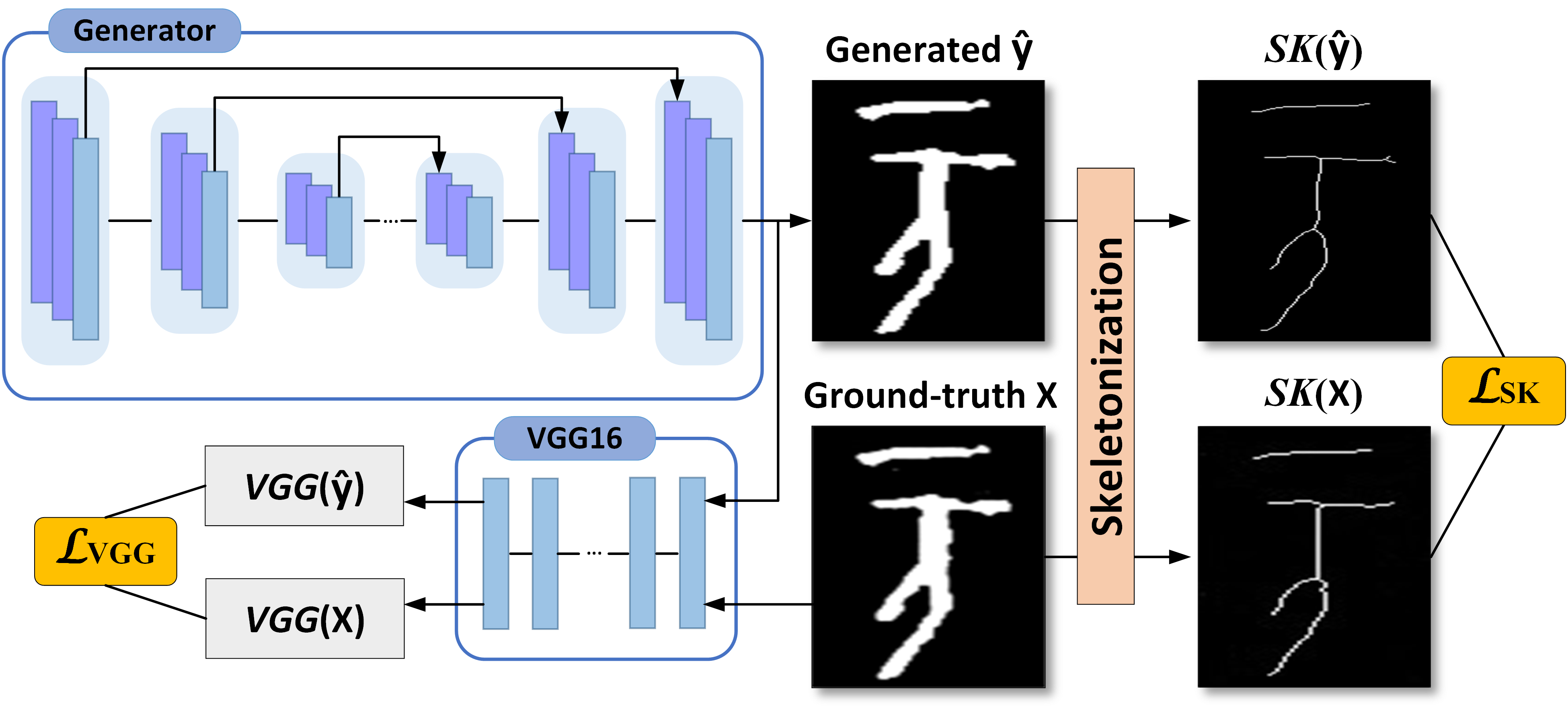}
	\caption{The architecture of generator. Skeleton loss and VGG loss are used for training the generator.
	\label{fig:4}}
\end{figure}

Several loss functions are applied for adjusting the training of the generator. We propose a novel pixel-based loss function $\mathcal{L}_{SK}$, namely skeleton loss, for enhancing model learning on skeleton information and keeping structural consistency of the restoration results. As shown in Figure~\ref{fig:4}, we obtain the skeleton of the ground-truth character image $x$ by our skeletonization method as $SK(x)$. Then we compare the skeleton $SK(x)$ with that of the generated image $SK(\hat{y})$. The skeleton loss function $\mathcal{L}_{SK}$ can be defined as: 
\begin{equation}
    \mathcal{L}_{SK} =  \frac{\theta_{SK}}{HW} * \sum_{h=0}^{H-1} \sum_{w=0}^{W-1} 	\left \| SK(\hat{y})_{h,w} - SK(x)_{h,w}  \right \|_1,
\label{equ:8}
\end{equation}
where $H$, $W$ are the height and width of the skeleton image, respectively, $h {\in} H$ and $w {\in} W$. 

Different from the skeleton loss calculated pixel-by-pixel, we also consider adjusting model training by feature-based loss. Thus, we apply VGG loss \cite{InvDN} aims to measure the global discrepancy between $x$ and $\hat{y}$. We exploit a pre-trained VGG16 \cite{46} to extract the feature maps of $x$ and $\hat{y}$. VGG loss can be presented as:
\begin{equation}
    \mathcal{L}_{VGG} = \theta_{VGG} \mathcal{L}_{MSE} (VGG(\hat{y}), VGG(x)),
\label{equ:9}
\end{equation}
where $\mathcal{L}_{MSE}(\cdot)$ is the loss function to calculate the averaged mean squared error (MSE) between the feature map $VGG(\hat{y})$ and $VGG(x)$. 

Meanwhile, we also apply the common pixel-wise image reconstruction loss $\mathcal{L}_{rec}$ and the GAN loss $\mathcal{L}_{GAN}^G$ to the generator, as follows: 
\begin{equation}
\mathcal{L}_{rec} = \theta_{rec} \left \| \hat{y} - x  \right \|_1,
\label{equ:11}
\end{equation}
\vspace{-8pt}
\begin{equation}
  \mathcal{L}_{GAN}^G = \theta_{GAN}log(1{-}D(\hat{y})),
\label{equ:10}
\end{equation}
Thus, we define the overall loss function for the generator $\mathcal{L}_{G}$ as:
\begin{equation}
    \mathcal{L}_{G} = \mathcal{L}_{GAN}^G + \mathcal{L}_{SK} + {L}_{VGG} + {L}_{rec},
\label{equ:10}
\end{equation}
where $\theta {=} \{\theta_{SK}, \theta_{VGG}, \theta_{GAN}, \theta_{rec}\}$ refers to the weights of the corresponding loss functions.

\noindent \textbf{Discriminator of CiRNet.}
The discriminator aims to distinguish generated images from real ones. Some GAN-based methods try to exploit global and local image-content consistency in their discriminator \cite{53}, where the global discriminator and the local discriminator focus on the inconsistency in the whole image and partial regions respectively. It will be beneficial for the discriminator if it can learn features in various scales simultaneously. Thus, we introduce SeR blocks into the discriminator for checking the image-content inconsistency under different scales, as shown in Figure~\ref{fig:2}. The discriminator consists of a convolutional layer as the input projector, five SeR blocks and a fully connected layer $FC$. The overall loss function of the discriminator is the GAN loss, which is given as follows: 
\begin{equation}
    \mathcal{L}_{D} = \mathcal{L}_{GAN}^D = - {\rm log}(D(r)) -  {\rm log}(1-D(\hat{y})),
\label{equ:11}
\end{equation}
where the function $D(\cdot)$ refers to the prediction result of the discriminator; $r$ stands for a real and clean image.

\section{Experiments}

\begin{figure*}[!t]
\setlength{\abovecaptionskip}{5pt}%
\setlength{\belowcaptionskip}{0pt}%
	\centering
	\includegraphics[width=0.95\linewidth]{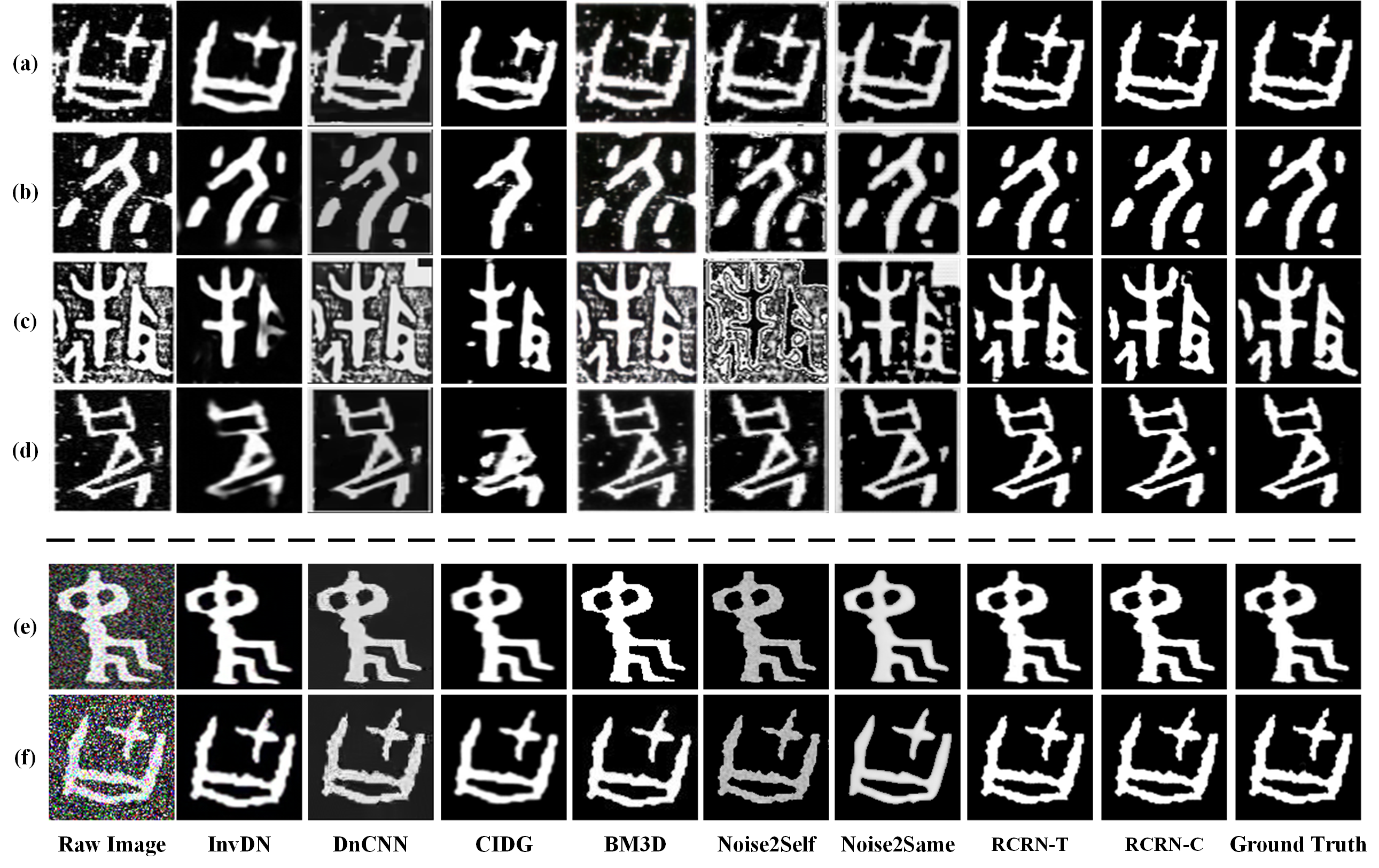}
    \caption{Comparison results of different methods on representative degraded character images. (a)-(d) restoration results on $D_{real}$; (e)-(f) restoration results on $D_{syn}$.
    \label{fig:5}}
\end{figure*}

\subsection{Experimental Setups}

\noindent\textbf{Datasets.}
Character image restoration, as a broad research topic, whose datasets contain diverse benchmarks, e.g., real-world/ synthetic noise, handwritten/scanned, and character/document. However, there is a lack of available datasets for real-world character image restoration, since existing publicly accessible datasets simply add synthetic noise to the character image. Therefore, we build a novel character image dataset\footnote{See link: https://github.com/daqians/Noisy-character-image-benchmark.} to provide valid benchmarks for recovering real-world degradation, namely $D_{real}$, by collecting from the historical Chinese character and oracle document datasets \cite{41}. The degradation model of these images follows Eq.~\eqref{equ:2}. The dataset includes training and testing sets consisting of 1467 and 139 noisy-clean character image pairs, respectively. Note that clean character images are binarized and are manually produced by several philologists. Moreover, to compare the restoration results in the images with real-world and synthetic noise for a more comprehensive evaluation, we generated another dataset called $D_{syn}$ by adding random Gaussian noise (noise variance $\sigma = [5,15]$) on our produced clean images.


\begin{table}[!t]
\centering
\caption{Quantitative evaluation results for RCRN and baseline methods on datasets $D_{real}$ and $D_{syn}$.The best and second-best results are emphasized in {\color[HTML]{FF0000} red} and {\color[HTML]{0000FF} blue}, respectively. }
\label{tab:1}
\begin{tabular}{@{}ccccc@{}}
\toprule
\multirow{2}{*}{Methods} & \multicolumn{2}{c}{$D_{real}$}   & \multicolumn{2}{c}{$D_{syn}$}    \\ \cmidrule(l){2-3}  \cmidrule(l){4-5} 
                         & PSNR $\uparrow$           & SSIM $\uparrow$            & PSNR $\uparrow$           & SSIM  $\uparrow$           \\ \midrule
Raw Image                & 7.82           & 0.4366          & 14.18          & 0.3406          \\
DnCNN                    & 11.45          & 0.5481          & 21.07          & 0.8683          \\
InvDN                    & 15.04          & 0.7138          & {\color[HTML]{FF0000} 22.86} &  0.8913 \\
CIDG                     & 14.83          & 0.6771          & 21.99         & {\color[HTML]{0000FF} 0.9060} \\
BM3D                     & 9.36           & 0.355           & 22.31           & 0.8176          \\
Noise2Self               & 11.04          & 0.5652          & 19.87           & 0.7910          \\
Noise2Same               & 13.08          & 0.5960          & 20.72          & 0.8794          \\ \midrule
RCRN-T     & {\color[HTML]{0000FF} 19.07} & {\color[HTML]{0000FF} 0.8129} & {\color[HTML]{0000FF} 22.52}  & 0.9037 \\
RCRN-C      & {\color[HTML]{FF0000} 19.81} & {\color[HTML]{FF0000} 0.8483} & 22.10          & {\color[HTML]{FF0000} 0.9095} \\
\bottomrule
\end{tabular}
\end{table}

\noindent\textbf{Baseline Methods.}
We consider several powerful image restoration methods as baselines to compare with our method, including InvDN \cite{InvDN}, calligraphic image denoising GAN (CIDG) \cite{14}, DnCNN \cite{16}, Noise2Self \cite{noise2self}, Noise2Same \cite{18}, and a classic denoising algorithm BM3D \cite{BM3D}. To fairly compare these methods, we train the former five methods on the dataset $D_{real}$ and $D_{syn}$, and apply the corresponding fixed settings to BM3D. Need to notice that Noise2Same, CIDG, Noise2Self are developed to handle degraded character images. The remaining methods are general image restoration methods that are widely applied in practice. 

\subsection{Quantitative Evaluation}
We quantitatively evaluate the methods by two commonly used metrics for low-level vision tasks, i.e., peak signal-to-noise ratio (PSNR) and structural similarity index measure (SSIM) \cite{54}. We involve metrics for raw character images to visually compare the performance. Table \ref{tab:1} shows the comparisons between our method and other baseline methods on datasets $D_{real}$ and $D_{syn}$. We introduce RCRN-T and RCRN-C which are developed based on SeR-T and SeR-C blocks, respectively. 

\noindent \textbf{Dataset $D_{real}$.} The PSNR/SSIM results indicate that the proposed RCRN-C significantly surpass all comparing methods when applying on $D_{real}$. DnCNN and BM3D poorly perform on $D_{real}$ since the noise in character images are much different from that in general images. Noise2Self and Noise2Same only achieve limited results although they have restored character images successfully in their experiments. This is because they mainly focus on synthetic noise while not designed for real-world noise. InvDN and CIDG are designed for the restoration of real-world noise and character images, also achieving promising performance. Benefiting from the skeleton information and powerful scale-ensemble image restorer, RCRN-C and RCRN-T achieve better results, which is clearly better than InvDN and CIDG.  

\noindent \textbf{Dataset $D_{syn}$.} Experimental results show all methods achieve better restoration performance on $D_{syn}$ than on $D_{real}$, indicating that synthetic degradation is easier to handle compared to complex real-world degradation. InvDN and our RCRN-C achieve the best PSNR and SSIM results, respectively. RCRN-T and CIDG also achieve promising results. It demonstrates that our proposed methods are effective on both datasets. Need to notice that the different performance of the remaining methods on $D_{real}$ and $D_{syn}$ also indicates that it is difficult and costly for character image restoration with real-world degradation.

\subsection{Qualitative Evaluation}
The qualitative results for dataset $D_{real}$ are presented in Figure \ref{fig:5}(a)-(d), visibly demonstrating the superiority of our RCRN compared with the listed methods. We can find that DnCNN and BM3D cannot effectively remove the noise in character images. Noise2Self and Noise2Same return better results, where parts of the images achieve good quality. However, there is a risk of failure when the noise level is increasing, such as Noise2Self shown in Figure \ref{fig:5}(c). Moreover, the above methods are not effective for large-scale noise, e.g., Figure \ref{fig:5}(a) and (c). Oppositely, InvDN and CIDG perform better on noise removal but lack preserving the character itself. For example, they remove too much useful information in Figure \ref{fig:5}(b) and (d), and destroy the structural consistency of the characters. We find that RCRN-T and RCRN-C can effectively remove the visible degradation and precisely keep the character. Benefiting from SENet and skeleton loss function, our method can preserve the character information even for seriously degraded character images, which is crucial for constructing high-quality image sets. Figure \ref{fig:5}(e)-(f) show the qualitative results of dataset $D_{syn}$, where all methods achieve better performance.

\subsection{Ablation Study}
We apply ablation studies to provide detailed evidence for proving each component in our proposed method is reasonable and effective.  

\begin{figure}[!t]
	\centering
	\includegraphics[width=0.95\linewidth]{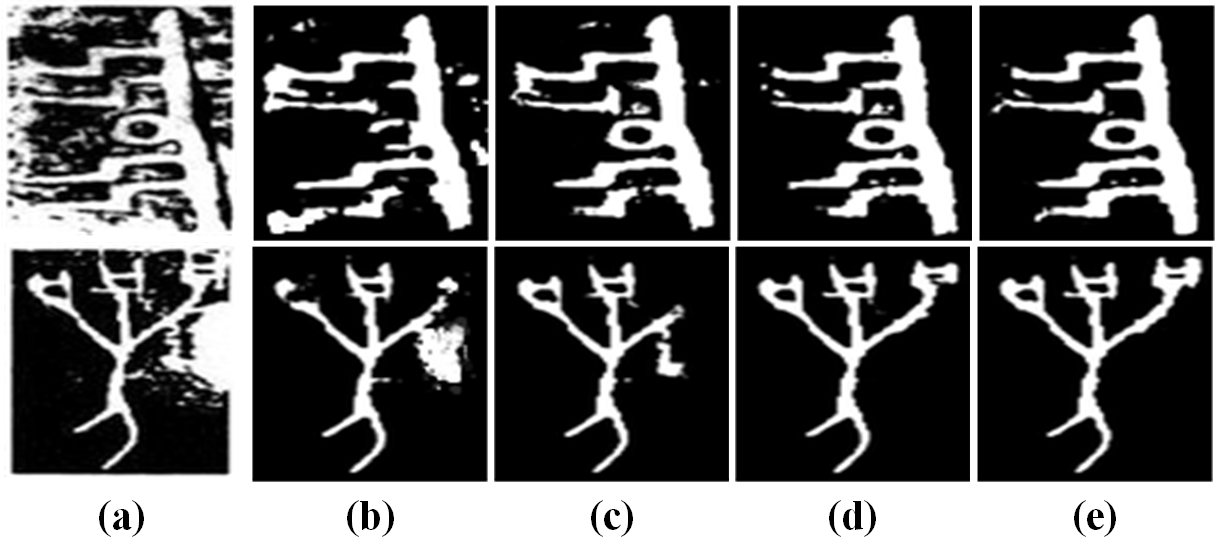}
    \caption{Qualitative results of ablation study. (a) raw image. (b) Backbone. (c) B+SeR-C (without SENet). (d) B+SeR-C (with single skeletonization). (e) RCRN-C.
	\label{fig:6}}
\end{figure}

\begin{table}[!t]
\centering
\captionsetup{font=small, labelfont=bf}
\setlength{\abovecaptionskip}{0pt}%
\setlength{\belowcaptionskip}{2pt}%
\caption{Ablation study results on CiRNet. }
\label{tab:2}
\resizebox{\columnwidth}{!}{
\begin{tabular}{@{}ccccccc@{}}
\toprule 
& \multicolumn{3}{c}{With SENet}   & \multicolumn{3}{c}{Without SENet}    \\ \cmidrule(l){2-4}  \cmidrule(l){5-7}
     & Backbone   & B+SeR-C & B+SeR-T & Backbone   & B+SeR-C & B+SeR-T  \\ \midrule
PSNR $\uparrow$  & 17.27                        & \textbf{19.81}                    & 19.07                    & 12.25                              & 14.04                & 12.95  \\
SSIM $\uparrow$  & 0.7302                       & \textbf{0.8483}                    & 0.8129                   & 0.5966                             & 0.7062               & 0.6133  \\ \bottomrule
\end{tabular}}
\end{table}

\noindent\textbf{Effect of CiRNet.} We propose an SeR layer in CiRNet for better restoring character images from mixed degradation. To verify its effectiveness, we remove the SeR layers and set a cGAN as the backbone (B) of CiRNet. We train and evaluate the models by clean-noisy character images pairs, with/without using skeletons. Table~\ref{tab:2} shows the PSNR/SSIM results and Figure~\ref{fig:6} shows two examples with mixed noise from $D_{real}$. It indicates that the mixed noise is removed by CiRNet to some extent. To be more specific, CiRNet achieves better quantitative results than the backbone.

\noindent\textbf{Effect of SENet.} In this experiment, we aim to present the effectiveness of restoring character images by skeleton extraction, more precisely, by SENet. We apply the proposed skeletonization in single \textit{+SK(ours)} and ensemble skeletonization \textit{+SENet}, and an additional skeletonization method \cite{chaussard2011robust} in single \textit{+SK\cite{chaussard2011robust}} for comparing with our methods, as shown in Table \ref{tab:3}. The CiRNet with SeR-C block is used as the backbone in this experiment. We can find that both PSNR and SSIM values improve a lot after applying skeletons. In Figure~\ref{fig:6}, we can find that the RCRN-C gives more stable restoration results where the character structure is preserved by SENet. Moreover, Table \ref{tab:2} presents the significant improvements of models with using SENet, which also proves the effectiveness of SENet.


\begin{table}[!t]
\centering
\captionsetup{font=small, labelfont=bf}
\setlength{\abovecaptionskip}{0pt}%
\setlength{\belowcaptionskip}{2pt}%
\caption{Ablation study results on SENet.}
\label{tab:3}
\resizebox{\columnwidth}{!}{
\begin{tabular}{@{}cccccc@{}}
\toprule
     & Backbone & B+SK(ours) & B+SK\cite{chaussard2011robust} & B+SENet \\ \midrule
PSNR/SSIM$\uparrow$ & 14.04/0.7062 & 17.49/0.7531 
&17.13/0.7509 & \textbf{19.81/0.8483} \\ \bottomrule
\end{tabular}}
\end{table}

\noindent\textbf{Effect of Loss Functions.} RCRN uses multiple loss functions, we apply a novel skeleton loss that aims to capture more structural information and a VGG loss to capture feature-level information. We compare the results of RCRN with/without these loss functions as shown in Table \ref{tab:4}, where the backbone B is RCRN-C using $\mathcal{L}_{rec} {+} \mathcal{L}_{GAN}^G$.

\begin{table}[!t]
\centering
\captionsetup{font=small, labelfont=bf}
\setlength{\abovecaptionskip}{0pt}%
\setlength{\belowcaptionskip}{2pt}%
\caption{Ablation study results on loss functions.}
\label{tab:4}
\resizebox{\columnwidth}{!}{
\begin{tabular}{@{}ccccc@{}}
\toprule
     & Backbone     & B+$\mathcal{L}_{VGG}$          & B+$\mathcal{L}_{SK}$           & B+$\mathcal{L}_{VGG}$  +$\mathcal{L}_{SK}$ \\ \midrule
PSNR/SSIM$\uparrow$ & 18.62/0.8036 & 19.03/0.8117 & 19.28/0.8132 & \textbf{19.81/0.8483} \\ \bottomrule
\end{tabular}}
\end{table}

\subsection{Application}
\begin{figure}[!t]
	\centering
	\includegraphics[width=0.8\linewidth]{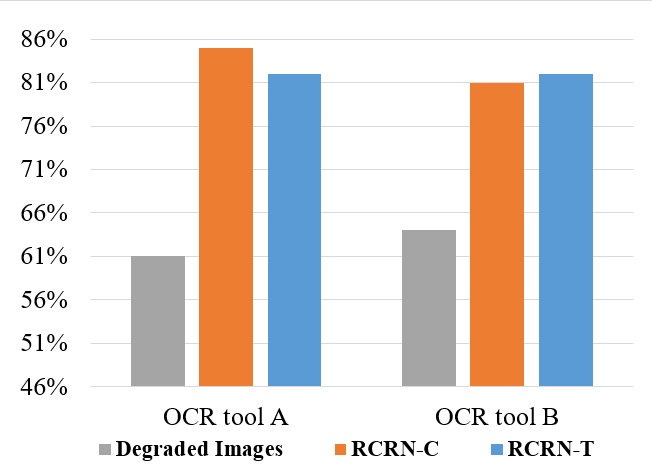}
	\caption{The recognition accuracy of degraded and restored character images by OCR tools.	
	\label{fig:7}}
\end{figure}

To further validate that our character image restoration method could be useful for computer vision applications and human recognition. We employ two optical character recognition (OCR) tools\footnote{https://www.jidagwz.com/ocr.html}$^,$\footnote{http://www.shufashibie.com/} of historical characters to validate whether RCRN can help to achieve better recognition performance. Figure~\ref{fig:7} shows the accuracy of the OCR results. We can find that the recognition results of restored character images significantly outperform that of the degraded character images (23\% and 17\% respectively), which can prove the proposed RCRN is instrumental for related applications by improving the quality of character images. 

\section{Conclusion}
We analyze complex degradation in real-world character images and clarify the issues for current image restoration methods. To solve these issues, we propose a novel framework, namely, RCRN, to effectively restore real-world degraded character images via skeleton extraction. In RCRN, character skeleton information is first-time applied to restore degraded characters since skeletons can preserve the structural consistency and improve the character reconstruction. We also propose a powerful image reconstruction network CiRNet by applying a scale-ensemble strategy. We compare RCRN with existing restoration methods on a newly collected dataset. Comparison results demonstrate the superiority of our method, in particular, for character images with real-world degradation. At the same time, the ablation study shows the effectiveness of each proposed improvement in our method. 

Meanwhile, RCRN can also be considered as a generic framework for computer vision tasks that take structural information as critical features. According to the promising evaluation results, we believe the proposed framework is adaptive to other character-based tasks, e.g., character style transfer and OCR. The potential applications also include human body/hand segmentation and action recognition, which will be our future study.

\begin{acks}

This research is supported by the National Natural Science Foundation of China (62077027), the program of China Scholarships Council (No.202007820024), the Ministry of Science and Technology of the People's Republic of China (2018YFC2002500), and the Department of Science and Technology of Jilin Province, China (20200801002GH).

\end{acks}

\bibliographystyle{ACM-Reference-Format}
\balance
\bibliography{full22}

\end{document}